\documentclass[10pt,twocolumn,letterpaper]{article}

\usepackage{iccv}
\usepackage{times}
\usepackage{epsfig}
\usepackage{graphicx}
\usepackage{graphics}
\usepackage{float}
\usepackage{amsmath}
\usepackage{amssymb}
\usepackage{multirow}

\newcommand{\N}{\mathbb{N}}

\usepackage[breaklinks=true,bookmarks=false]{hyperref}

\iccvfinalcopy 

\def\iccvPaperID{8} 

\ificcvfinal\fi

\begin{document}

\title{Soft Expectation and Deep Maximization for Image Feature Detection}

\author{Alexander Mai \\
University of California, San Diego \\
9500 Gilman Dr, La Jolla, CA 92093 \\
{\tt\small atm008@ucsd.edu}
\and
Allen Yang \\
University of California, Berkeley \\
Berkeley, CA \\
{\tt\small yang@eecs.berkeley.edu}
\and 
Dominique E. Meyer \\
University of California, San Diego \\
9500 Gilman Dr, La Jolla, CA 92093 \\
{\tt\small dom@ucsd.edu}
}

\maketitle
\ificcvfinal\thispagestyle{empty}\fi

\begin{abstract}
Central to the application of many multi-view geometry algorithms is the extraction of matching points between multiple viewpoints, enabling classical tasks such as camera pose estimation and 3D reconstruction. 
Many approaches that characterize these points have been proposed based on hand-tuned appearance models or data-driven learning methods. We propose Soft Expectation and Deep Maximization (SEDM), an iterative unsupervised learning process that directly optimizes the repeatability of the features by posing the problem in a similar way to expectation maximization (EM). We found convergence to be reliable and the new model to be more lighting invariant and better at localize the underlying 3D points in a scene, improving SfM quality when compared to other state of the art deep learning detectors.
\end{abstract}

\section{Introduction}\label{introduction}
A sparse image feature is a region in image space that corresponds to an object location in world space that can be reliably found and matched across multiple viewpoints. Specifically, our paper looks at keypoints, also known as corner or point features, which are elements $p\in\mathbb{R}^2$ in image space, with the goal of forming point correspondence between multiple images for use in various problems such as finding the relative pose between two perspective images, Structure-from-Motion (SfM), Simultaneous Localization and Mapping (SLAM), and many more \cite{multiviewgeometry}.

\begin{figure}
	\centering
    \includegraphics[width=\linewidth]{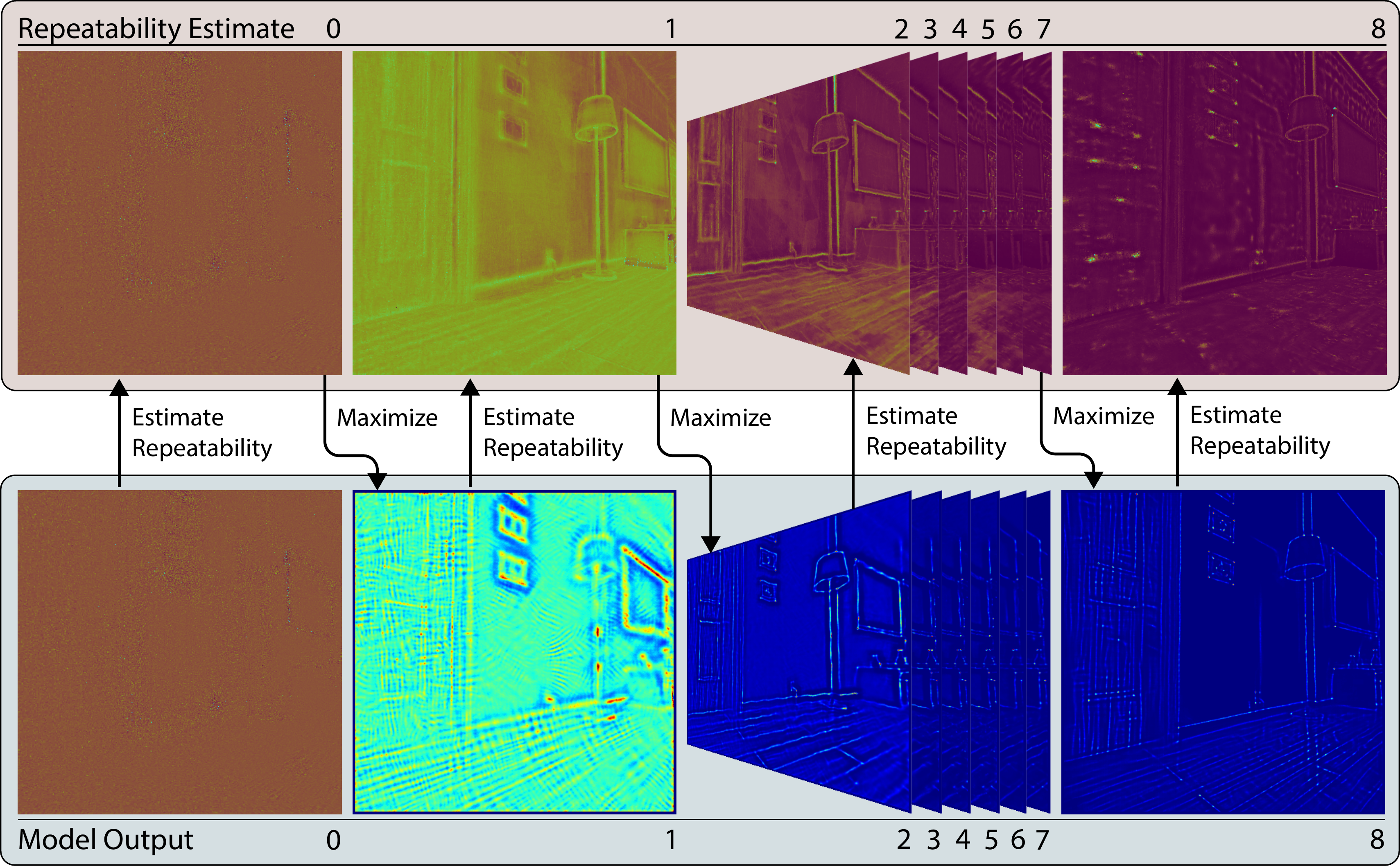}
    \caption{The top row depicts the results for the expectation step, mapping pixels to repeatability values. The bottom row shows the output of the network on the same images after training the network to maximize the repeatability values.}\label{fig:frontpage}
  \end{figure}

Recently, deep learning has been successfully applied to many vision applications, so the question is whether feature detection can be improved as well. One potential application of improved detection would be to augmented reality, where precise tracking is needed to align the virtual and real world.
However, feature detection is not a case of simply finding the right model to fit the data. 
One of the most important attributes of a feature detector is it's repeatability, or the probability that the detector will assign a 3D point a feature given that it has assigned that point a feature in a different viewpoint or under different lighting.
As such, the repeatability of any point is more so a property of the detector rather than the data.

To solve this problem, we formulate it as one of optimization, where the objective is to optimize the repeatability of the features in 3D space.
Since the problem is not differentiable, we find a solution to the optimization using an Expectation Maximization algorithm. 
We call our formulation \emph{Soft Expectation, Deep Maximization}, or SEDM for short.
In the expectation step, we estimate the repeatability of potential ground truth features using the current model parameters.
In the maximization step, we find features that maximize those repeatability estimates and use them as pseudo ground truth to train a new model. 
We find that SEDM is able to improve feature detection when compared to previous state of the art deep learning methods, even when trained on a small synthetic dataset.

Our main contributions are summarized as follows:
\begin{itemize}
\item We introduce SEDM, a novel method for training an image feature detector by optimizing the repeatability of corresponding features in 3D space.
\item We show it is possible to bridge the synthetic to real gap solely by using photorealistic images.
\item Our benchmarks provide evidence that SEDM is able to improve feature detection accuracy when compared to previous state of the art deep learning methods.
\item Our method achieves state of the art performance under illumination change.
\end{itemize}

\section{Related Work}
Methods for detecting image features are typically split into two categories: traditional and deep learning. This split aligns with the difference in the number of parameters to tune, ranging from traditional model-based methods having a few dozen to deep learning methods with millions. FAST \cite{fast}, which used machine learning to build a decision tree to classify whether a point was a corner on a dataset labeled using an ad hoc method, is still considered a traditional method by this definition. Of the traditional methods, SIFT \cite{sift} remains a gold standard in SfM reconstruction accuracy \cite{schonberger2017comparative}. When combined with an improved descriptor like HardNet \cite{hardnet} and a guided matcher, SIFT still beats other state-of-the-art methods on leaderboards like the Image Matching Challenge 2020 for 8k features \cite{imw2020}.

\subparagraph{Deep Learning in Feature Detection.}

One of the most successful methods, Superpoint \cite{superpoint} , has proven very capable, especially when combined with SuperGlue matcher \cite{superglue}. This combination has won several benchmarks, including the Image Matching Challenge for 2048 features \cite{imw2020} and the Visual Localization Benchmark on several datasets \cite{visualizationbenchmark}. SuperPoint differs from most other approaches in that it starts with ground-truth 3D geometry with known features. To train MagicPoint, the supervisor of SuperPoint, the method generated a synthetic dataset of simple lines and geometric shapes and labeled the corners as features. MagicPoint is used to supervise SuperPoint so it can learn to predict these same corners under image warping. The process by which SuperPoint is trained, known as iterative homographic adaptation, is also similar to our own. During each iteration, the output of the previous model is merged between an image and it's homographically warped version to produce the pseudo ground truth used to train the new model. Our method can be thought of an extension of homographic adaptation to 3D space. However, rather than starting with the trained MagicPoint detector, we start from scratch.

\begin{figure*}[ht!]
\begin{center}
\includegraphics[width=0.9\linewidth]
                   {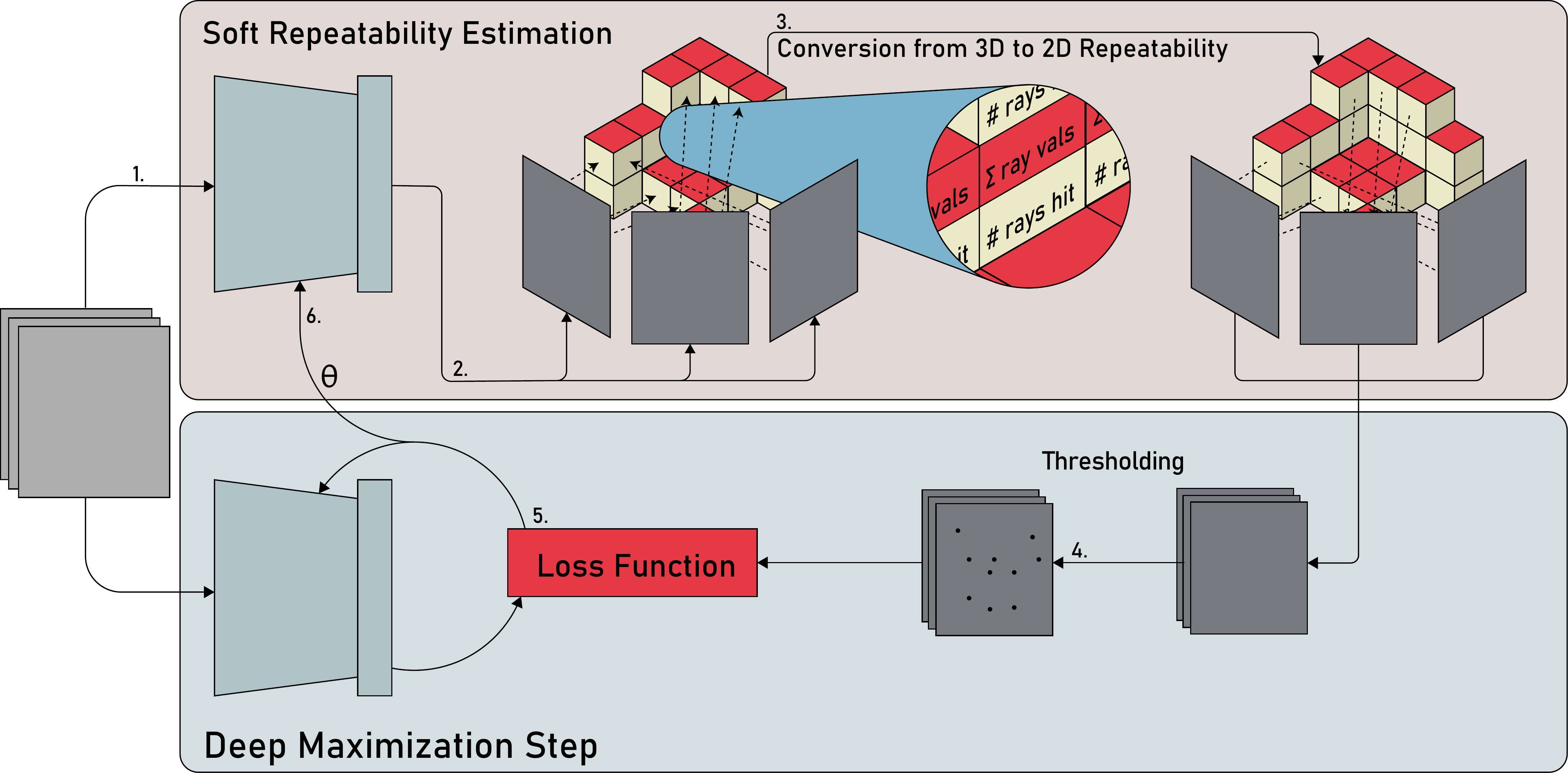}
\end{center}
   \caption{We start by initializing the network (depicted twice) with random weights. 1. We perform inference on images to generate heatmaps for the location of features in image space. 2. Project heatmap values onto the known voxel map of scene and accumulate the average heatmap value across all viewpoints within all voxels, as seen in the zoom in. 3. We project these soft 3D repeatability values back into image space. 4. We then use these soft 2D repeatability values to estimate the optimal features to detect for each image. 5. We minimize the loss function for the model using these estimates as a pseudo ground truth. 6. Lastly, we use these new model parameters to repeat the process from step 1.}
\label{fig:diagram}
\end{figure*}

Using image warping to approximate viewpoint change is a common method for optimizing towards repeatability between two or more heatmaps, such as in Key.Net  \cite{keynet}, Unsuperpoint \cite{unsuperpoint}, DetNet  \cite{lenc}. This method can be extended to model nonplanar change of perspective by extracting image patches around keypoints and warping the individual patches, such as in QuadNet \cite{quadnet} and R2D2 \cite{r2d2} . SIPs \cite{sips} utilizes point tracking and a sparse loss to perform what could be thought of as a sparse warp. These methods utilize some kind of approximation of projection and reprojection, which likely hinders the accuracy of the detector. LF-Net\cite{lfnet} utilizes the depth to directly back project one heatmap into the space of another, but limit the viewpoint change by only selecting pairs that are temporally adjacent.  
TILDE \cite{tilde} optimizes towards repeatability without warping heatmaps by learning on stationary webcams.
LIFT \cite{lift} filters keypoints output by SIFT using SfM and treats those as ground truth. 
D2Net \cite{d2net} and ASLFeat \cite{aslfeat} attempt to circumvent learning feature detection directly by extracting feature locations out of the feature map used to calculate descriptors.
DISK \cite{disk} and Reinforced Feature Point \cite{rfp} utilize reinforcement learning, with DISK using depth and pose to check whether matches are valid without any keypoints localization loss and Reinforced Feature Points only using the pose to evaluate the estimate.

\subparagraph{Expectation Maximization in Deep Learning.}
Expectation Maximization (EM) is not new to deep learning. 
Multiple methods have introduced random sampling into their methods, allowing sampling the output distribution and performing something similar to EM. 
One such method is variational autoencoders \cite{kingma2013auto}, whose optimization can be formulated as an EM algorithm where the latent variable is the learned embedding \cite{mcallester_2017}. 
Another instance is Bayesian neural networks \cite{jospin2020hands}, for which applying the EM algorithm is straightforward. 

At a broader level, there are methods similar to ours that do not fall under the label of EM. For methods that cannot output "well-calibrated" probabilistic predictions, there exist similar approaches under the label "wrapper" methods by taxonomists \cite{van2020survey}. Wrapper methods are model agnostic methods that alternate between training classifiers on pseudo-labeled data, then use the predictions of the resulting classifier to predict new pseudo-labels.

\section{SEDM Framework}


During model training, the process we call SEDM, we take as input: images of a scene, with each element of the scene viewed from different angles, a voxel map of the scene, and known poses for each image. During inference, the trained model, which we call the SEDM detector, takes an image as input and outputs a heatmap of equal width and height to the image, which is thresholded to obtain a list of feature locations in the image. These can then be paired with various descriptors and matchers for traditional feature pipelines.

The goal of SEDM is to maximize the repeatability of the features predicted by the SEDM detector. 
In the first step of the iterative process, the expectation step, we find the (soft) expected value of the repeatability of the features predicted by the SEDM detector by utilizing the voxel map and known poses of each image.
In the second step, the maximization step, we find a set of pseudo ground truth 3D features that maximize our estimate of the repeatability values and fit our model to that set. By iterating over these steps, we improve the quality of our detector, as seen in Figure \ref{fig:frontpage}.

\subsection{Soft Expectation Step}\label{modeleval}

To maximize the repeatability of the SEDM detector, we start by estimating the repeatability of the features it produces.
One way to do this would be to perform 3D reconstruction on the scene using the features predicted by the SEDM detector, calculate the reprojection error for each predicted feature, then estimate the repeatability for some reprojection error threshold empirically across the viewpoints in the scene. However, if instead we use datasets for which we have a model of the environment being photographed and the exact pose of each image, we can perform a much more precise estimate of the repeatability of each feature. 
Instead of reconstructing the scene and finding the repeatability for each point, we represent the space of possible features as 3D voxels, find the repeatability of each of the feature voxel, then project all of the estimates back into image space. This gives rise to a reprojection error threshold in 3D space equal to the extent of each voxel (0.005m in our experiments).

For each viewpoint image and pixel in that image, we raycast into the scene and increment two counters on each voxel; one counts whether the SEDM detector detected a feature at that pixel, denoted as $D$ and the other counts the number of times a ray hit that voxel, denoted as $N$. The expected value of the repeatability of that feature is $D/N$.
However, rather than selecting a threshold for what scores counted as a detection, we find a soft repeatability value by taking a sum over the score assigned by the SEDM detector rather than the indicator function of that score. Once we have this soft repeatability value for each voxel in 3D space, we backproject it into 2D image space.

\subsection{Deep Maximization Step\label{prediction}}
We want to train the SEDM detector to predict features that would maximize the expected repeatability using our estimates. To do so, we find a set of pseudo ground truth features that maximize the expected repeatability values, then fit the model to the pseudo ground truth using cross entropy loss between the model heatmap and binary mask depicting feature locations:
\[L(x, y) = -\sum_{i=1}^n\sum_{j=1}^m y_{ij}\log x_{ij} + (1-y_{ij})\log(1-x_{ij})\]

When this optimization is unconstrained, the set of features in 2D that maximizes these expected repeatability values is actually the set of all pixels in the image. Since detecting every pixel as a feature isn't very useful, we apply a few constraints to the set of features being maximized. We limit the number of features in each image to $L\in\N$ and require that the points be least some distance $r_{\text{nms}}$ from each other.

The constraint that there be a finite number of points keypoints per image was essentially implemented as a shadow price whose weight was controlled by the maximum number of ground truth keypoints per an image. As we iterated the algorithm, we reduced this number, stopping when important keypoints started disappearing. However, this method alone was prone to local minima where edges were prioritized over corners. Filtering edges from the psuedoground truth was enough to close the duality gap and ensure reliable convergence.

\begin{figure}[t]
\begin{center}
\includegraphics[width=0.95\linewidth]{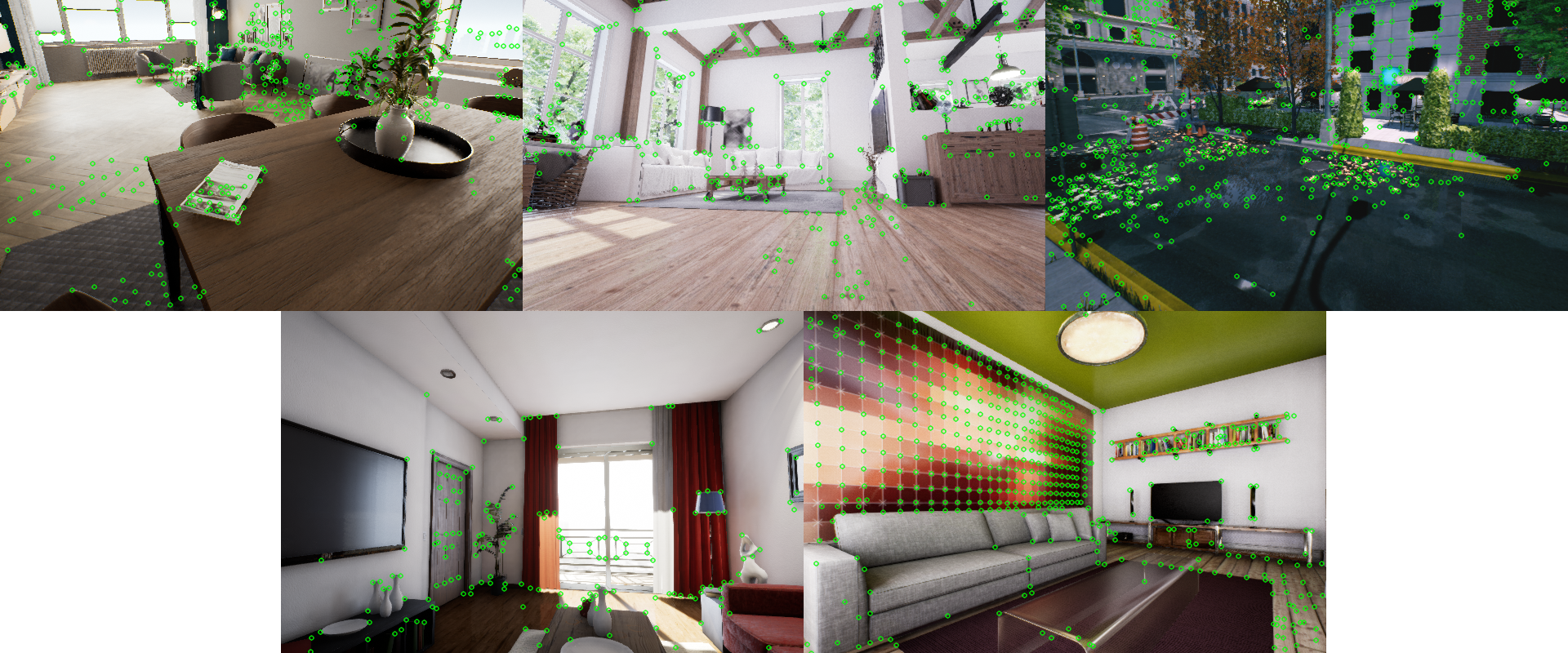}
   \caption{The synthetic datasets used for training. From left to right: ArchVisInterior, ArchinteriorsVol2Scene1, UrbanCity, RealisticRendering, ArchinteriorsVol2Scene2. The green dots are the SEDM detector features.}
\label{fig:hpatches}
\end{center}
\end{figure}

\section{Implementation}
\subsection{Training}
We used SuperPoint \cite{superpoint} as the SEDM detector model architecture. 
To review, the SuperPoint architecture consists of a VGG-style encoder followed by two decoders. One decoder consists of a few extra convolution layers to output dense descriptors across the image, which can be sampled using bi-cubic interpolation to output descriptors for sparse features. The second decoder consists of a few extra convolution layers, followed by a depth wise softmax and a depth to space encoder that scales the output heatmap back up to the original image size.
We added batch normalization layers\cite{ioffe2015batch} after the convolution layers and trained the model from scratch using PyTorch \cite{pytorch}. For each iteration, SuperPoint was trained for 300 epochs with a batch size of 32 using the Adam \cite{kingma2014adam} optimizer, $lr=0.0001, \beta=(0.9, 0.999)$. Images were augmented using the ImgAug \cite{imgaug} library to apply random Gaussian blur, salt and pepper noise, random linear contrast, additive Gaussian noise, brightness variation, JPEG compression, and random affine and perspective transformations. There are a couple miscellaneous necessary parameters and operations. A 4 pixel border around the heatmap was omitted during both training and testing. NMS radius $r_{nms}$ was 6 pixels during training and 3 during testing. Cellwise suppression method, where only one keypoint was selected in each $8\times 8$ cell was applied to match the softmax cells in the output of SuperPoint. The number of desired keypoints $L$ was annealed every third iteration to follow the sequence [2000, 1700, 1200]. 

During testing, the output threshold was set to $0.025$ and we paired the SEDM detector output with descriptors from a pretrained SuperPoint model and matched the descriptors using nearest neighbor matching. We provide the source code at \url{https://gitlab.com/confidence_painting/cpaint}.

\subsection{System Runtime}
When using SuperPoint descriptors, we must run the same model architecture twice, once with our weights and once with the original author's. According to \cite{superpoint}, this gives us a runtime of 23.65 ms, or 42 FPS on a Titan X GPU at a resolution of $480\times 640$.

\subsection{Training Dataset}

SEDM requires that the training dataset have precise, dense depth that aligns with the visual images, as well as accurate pose. Each element of the scene must be captured from different angles, otherwise that element is masked during training.
By leveraging the UnrealCV plugin \cite{qiu2017unrealcv} for the Unreal Engine, we were able to manually tour the camera around the scenes to capture this data.
We used the following environments: RealisticRendering and Archviz Interior by Epic Games, Archinteriors Vol 2 by Evermotion, and Urban City by PolyPixel.
We also selected seven high quality scenes from the BlendedMVS \cite{blendedmvs} dataset, which renders photorealistic images from an SfM model. Together, these two datasets totaled 45k images at $480\times 640$ resolution.


\begin{table}[]
    \centering
    \begin{tabular}{|c|c|c|c|}
        \hline
        Method & Illumination & Homography & Mean \\
        \hline
         & \multicolumn{3}{c|}{MMA @ $<$1 pixel err} \\
        \hline
        SEDM Detector & \textbf{45.3\%} & 30.1\% & \textbf{37.4\%} \\
        SuperPoint & 40.8\% & 30.1\% & 35.2\% \\
        SIFT & 36.5\% & \textbf{32.8\%} & 34.6\% \\
        R2D2 & 32.3\% & 22.8\% & 27.4\% \\
        \hline
         & \multicolumn{3}{c|}{MMA @ $<$2 pixel err} \\
        \hline
        SEDM Detector & \textbf{62.3\%} & 51.0\% & 56.4\% \\
        SuperPoint & 60.6\% & 53.1\% & \textbf{56.7\%} \\
        SIFT & 45.0\% & 46.7\% & 45.9\% \\
        R2D2 & 54.4\% & \textbf{58.0\%} & 54.4\% \\
        \hline
    \end{tabular}
    \caption{Results on the HPatches repeatability benchmark. Mean matching accuracy (MMA) at different maximum reprojection error thresholds for which a pair of points would be considered a match. Our SEDM detector is especially invariant to illumination change, possibly due to our ability to simulation illumination change in our synthetic environments.}
    \label{table:hpatches}
\end{table}
\section{Experiments}
\subsection{Image Matching}
In the HPatches Mean Matching Accuracy (MMA) test, the first image is matched against all others in the sequence, and the reprojection error is obtained using the known homography between pairs \cite{d2net}\cite{hpatches}. The number of pairs with error less than a varying threshold is then used to obtain an accuracy score.
The HPatches dataset has two sequences of images: illumination and viewpoint (homography).
For the illumination sequences, images from stationary cameras throughout the day are compared.
The viewpoint test has its limitations, as the planar viewpoint changes make the results hard to compare to the real world. 


The results are shown in Table \ref{table:hpatches}. At the most important error threshold, 1 pixel of error, the SEDM detector outperforms all other methods we compared against. 
This is especially true of illumination, where our ability to utilize photorealistic lighting changes within each scene seems to have led to an improvement in performance. Since both SIFT and SuperPoint utilize homographies to represent viewpoint change and our method does not, it is not surprising to see those methods outperform ours in that category.

\subsection{3D Reconstruction}
\begin{figure}[ht!]
\begin{center}
\includegraphics[height=130pt]{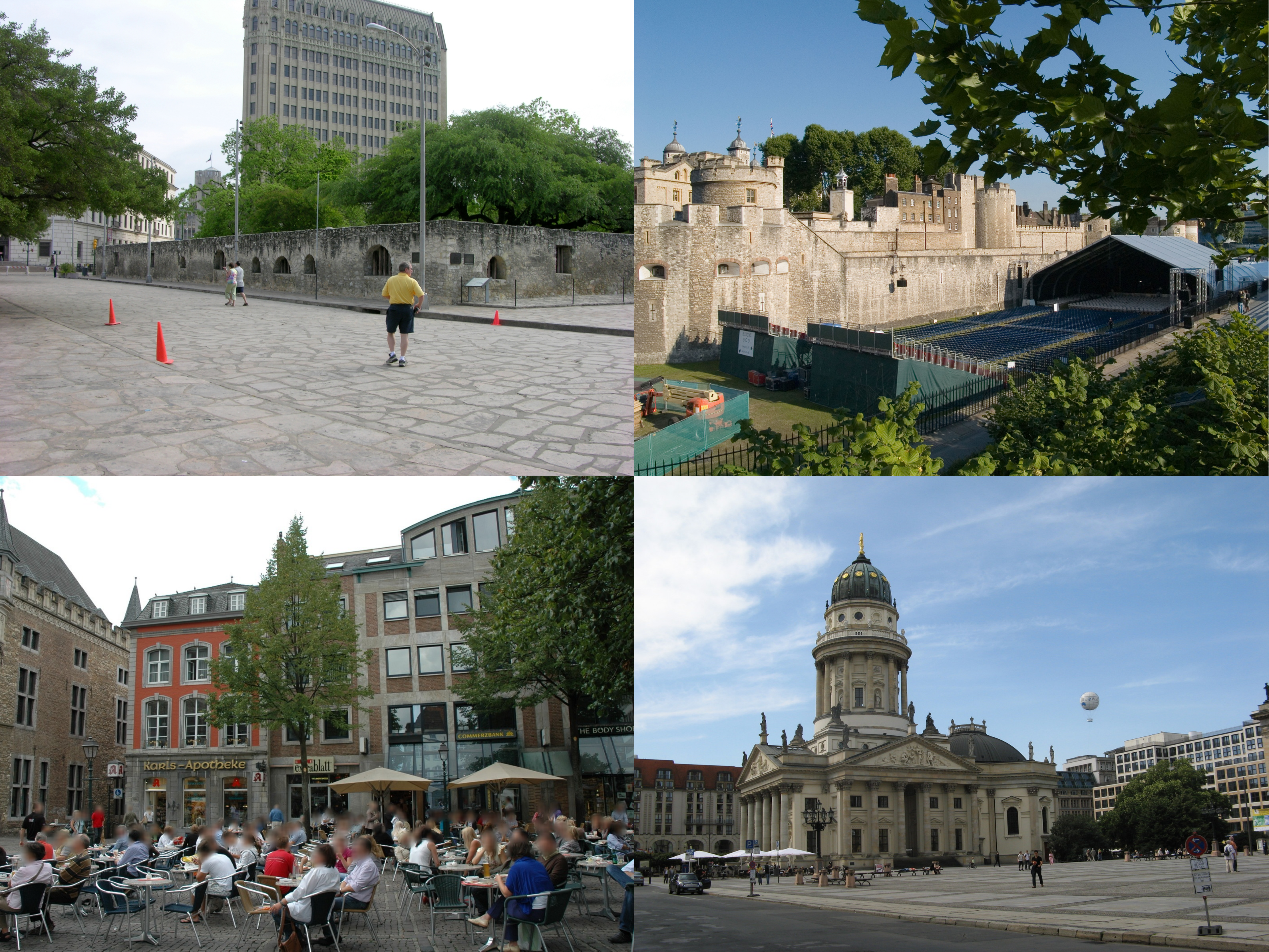}
   \caption{The real world datasets used for testing 3D reconstruction. From left to right: Alamo, Tower of London, Aachen Day Night, and Gendarmenmarkt.}
\label{fig:reconstruction}
\end{center}
\end{figure}
As in the survey by Schonburger et al. \cite{schonberger2017comparative}, we use COLMAP \cite{schonberger2016sfm} to perform sparse 3D reconstruction and report the statistics for the reconstruction, focusing on mean reprojection error. The mean reprojection error is the residual error after minimization between the predicted detection location and the reprojected 3D point in image space. We performed reconstruction for Alamo, Gendarmenmarkt, and Tower of London \cite{wilson_eccv2014_1dsfm}, and Aachen Day Night \cite{sattler2012image} to test outdoor performance. We also performed separate tests on our own training sets to test the effect of the synthetic to real jump. 

We perform a two tailed paired T-test at $\alpha=0.05$ on the mean difference between the reprojection error of our method versus SIFT, SuperPoint, and R2D2, with each dataset constituting a data point. A 95\% confidence interval for the difference between our method and these other methods is shown in Table \ref{table:reconstruction}. As can be seen, our method outperforms other machine learning methods. Just like in the original survey  \cite{schonberger2017comparative}, SIFT still outperforms all other methods on real world datasets. However, on the sythetic datasets, the difference between the SEDM detector and SIFT becomes statistically insignficant, implying a possible sythetic to real gap. Another possible cause is that most of the synthetic datasets are indoors while all the real world datasets are outdoors.
\begin{table}[t]
\begin{center}
  \begin{tabular}{c|c|c}
      \multicolumn{3}{c}{Mean Reprojection Error Diff vs SEDM Detector} \\
      \hline
      \hline
      Method & Real & Synthetic \\
      \hline
      Superpoint & $0.15\pm0.11\text{px}$ & $0.20\pm0.03\text{px}$ \\
      R2D2 & $0.38\pm0.23 \text{px}$ & $0.20\pm0.17 \text{px}$\\
      SIFT & $-0.30\pm0.09 \text{px}$ &  $-0.04\pm0.15 \text{px}$ \\
  \end{tabular}
   \caption{Results on our 3D reconstruction benchmark using COLMAP. The mean difference in mean reprojection error between our SEDM detector and other methods is shown. Lower is better.}
\label{table:reconstruction}
\end{center}
\end{table}

\section{Discussion and Conclusion}
We have presented SEDM, a novel method that allows us to optimize for keypoint repeatability in 3D space.
Rather than working within image space, SEDM enables us to work with features within the 3D space that the camera images.
We believe that this formulation for training keypoint detectors allows for more precise control over desired feature characteristics.
We consider our method to be a minimal implementation of SEDM, and yet our experiments demonstrate that the resulting detector achieves state of the art results, achieving best performance against illumination change on the HPatches benchmark and best performance among deep learning detectors in 3D reconstruction.

{\small
\bibliographystyle{ieee_fullname}
\bibliography{egbib}

\begin{thebibliography}{10}\itemsep=-1pt

\bibitem{multiviewgeometry}
Alex~M Andrew.
\newblock Multiple view geometry in computer vision.
\newblock {\em Kybernetes}, 2001.

\bibitem{hpatches}
Vassileios Balntas, Karel Lenc, Andrea Vedaldi, and Krystian Mikolajczyk.
\newblock {HPatches:} a benchmark and evaluation of handcrafted and learned
  local descriptors.
\newblock In {\em Proceedings of the IEEE conference on computer vision and
  pattern recognition}, pages 5173--5182, 2017.

\bibitem{keynet}
Axel Barroso-Laguna, Edgar Riba, Daniel Ponsa, and Krystian Mikolajczyk.
\newblock {Key.Net:} keypoint detection by handcrafted and learned cnn filters.
\newblock In {\em Proceedings of the IEEE International Conference on Computer
  Vision}, pages 5836--5844, 2019.

\bibitem{rfp}
Aritra Bhowmik, Stefan Gumhold, Carsten Rother, and Eric Brachmann.
\newblock Reinforced feature points: Optimizing feature detection and
  description for a high-level task.
\newblock In {\em Proceedings of the IEEE/CVF conference on computer vision and
  pattern recognition}, pages 4948--4957, 2020.

\bibitem{unsuperpoint}
Peter~Hviid Christiansen, Mikkel~Fly Kragh, Yury Brodskiy, and Henrik Karstoft.
\newblock Unsuperpoint: End-to-end unsupervised interest point detector and
  descriptor.
\newblock {\em arXiv preprint arXiv:1907.04011}, 2019.

\bibitem{sips}
Titus Cieslewski, Konstantinos~G Derpanis, and Davide Scaramuzza.
\newblock Sips: Succinct interest points from unsupervised inlierness
  probability learning.
\newblock In {\em 2019 International Conference on 3D Vision (3DV)}, pages
  604--613. IEEE, 2019.

\bibitem{superpoint}
Daniel DeTone, Tomasz Malisiewicz, and Andrew Rabinovich.
\newblock Superpoint: Self-supervised interest point detection and description.
\newblock In {\em Proceedings of the IEEE conference on computer vision and
  pattern recognition workshops}, pages 224--236, 2018.

\bibitem{d2net}
Mihai Dusmanu, Ignacio Rocco, Tomas Pajdla, Marc Pollefeys, Josef Sivic,
  Akihiko Torii, and Torsten Sattler.
\newblock {D2-Net:} a trainable cnn for joint detection and description of
  local features.
\newblock {\em arXiv preprint arXiv:1905.03561}, 2019.

\bibitem{ioffe2015batch}
Sergey Ioffe and Christian Szegedy.
\newblock Batch normalization: Accelerating deep network training by reducing
  internal covariate shift.
\newblock In {\em International conference on machine learning}, pages
  448--456. PMLR, 2015.

\bibitem{imw2020}
Yuhe Jin, Dmytro Mishkin, Anastasiia Mishchuk, Jiri Matas, Pascal Fua,
  Kwang~Moo Yi, and Eduard Trulls.
\newblock Image matching across wide baselines: From paper to practice.
\newblock {\em International Journal of Computer Vision}, pages 1--31, 2020.

\bibitem{jospin2020hands}
Laurent~Valentin Jospin, Wray Buntine, Farid Boussaid, Hamid Laga, and Mohammed
  Bennamoun.
\newblock Hands-on bayesian neural networks--a tutorial for deep learning
  users.
\newblock {\em arXiv preprint arXiv:2007.06823}, 2020.

\bibitem{imgaug}
Alexander~B. Jung.
\newblock {imgaug}.
\newblock \url{https://github.com/aleju/imgaug}, 2018.
\newblock [Online; accessed 30-Oct-2018].

\bibitem{kingma2014adam}
Diederik~P Kingma and Jimmy Ba.
\newblock Adam: A method for stochastic optimization.
\newblock {\em arXiv preprint arXiv:1412.6980}, 2014.

\bibitem{kingma2013auto}
Diederik~P Kingma and Max Welling.
\newblock Auto-encoding variational bayes.
\newblock {\em arXiv preprint arXiv:1312.6114}, 2013.

\bibitem{lenc}
Karel Lenc and Andrea Vedaldi.
\newblock Learning covariant feature detectors.
\newblock In {\em European conference on computer vision}, pages 100--117.
  Springer, 2016.

\bibitem{sift}
David~G Lowe.
\newblock Distinctive image features from scale-invariant keypoints.
\newblock {\em International journal of computer vision}, 60(2):91--110, 2004.

\bibitem{aslfeat}
Zixin Luo, Lei Zhou, Xuyang Bai, Hongkai Chen, Jiahui Zhang, Yao Yao, Shiwei
  Li, Tian Fang, and Long Quan.
\newblock {ASLFeat:} learning local features of accurate shape and
  localization.
\newblock In {\em Proceedings of the IEEE/CVF Conference on Computer Vision and
  Pattern Recognition}, pages 6589--6598, 2020.

\bibitem{mcallester_2017}
David McAllester.
\newblock {VAE = EM}.
\newblock Machine Thoughts.
\newblock https://machinethoughts.wordpress.com/2017/10/02/vae-em/.
\newblock (Accessed Mar 2021).

\bibitem{hardnet}
Anastasiya Mishchuk, Dmytro Mishkin, Filip Radenovic, and Jiri Matas.
\newblock Working hard to know your neighbor's margins: Local descriptor
  learning loss.
\newblock {\em arXiv preprint arXiv:1705.10872}, 2017.

\bibitem{lfnet}
Yuki Ono, Eduard Trulls, Pascal Fua, and Kwang~Moo Yi.
\newblock Lf-net: Learning local features from images.
\newblock {\em arXiv preprint arXiv:1805.09662}, 2018.

\bibitem{pytorch}
Adam Paszke, Sam Gross, Francisco Massa, Adam Lerer, James Bradbury, Gregory
  Chanan, Trevor Killeen, Zeming Lin, Natalia Gimelshein, Luca Antiga, et~al.
\newblock Pytorch: An imperative style, high-performance deep learning library.
\newblock {\em Advances in neural information processing systems},
  32:8026--8037, 2019.

\bibitem{qiu2017unrealcv}
Weichao Qiu, Fangwei Zhong, Yi Zhang, Siyuan Qiao, Zihao Xiao, Tae~Soo Kim,
  Yizhou Wang, and Alan Yuille.
\newblock {UnrealCV:} virtual worlds for computer vision.
\newblock {\em ACM Multimedia Open Source Software Competition}, 2017.

\bibitem{r2d2}
Jerome Revaud, Philippe Weinzaepfel, C{\'e}sar De~Souza, Noe Pion, Gabriela
  Csurka, Yohann Cabon, and Martin Humenberger.
\newblock R2d2: repeatable and reliable detector and descriptor.
\newblock {\em arXiv preprint arXiv:1906.06195}, 2019.

\bibitem{fast}
Edward Rosten and Tom Drummond.
\newblock Machine learning for high-speed corner detection.
\newblock In {\em European conference on computer vision}, pages 430--443.
  Springer, 2006.

\bibitem{superglue}
Paul-Edouard Sarlin, Daniel DeTone, Tomasz Malisiewicz, and Andrew Rabinovich.
\newblock {SuperGlue:} learning feature matching with graph neural networks.
\newblock {\em 2020 IEEE/CVF Conference on Computer Vision and Pattern
  Recognition (CVPR)}, pages 4937--4946, 2020.

\bibitem{visualizationbenchmark}
Torsten Sattler, Will Maddern, Carl Toft, Akihiko Torii, Lars Hammarstrand,
  Erik Stenborg, Daniel Safari, Masatoshi Okutomi, Marc Pollefeys, Josef Sivic,
  et~al.
\newblock Benchmarking 6dof outdoor visual localization in changing conditions.
\newblock In {\em Proceedings of the IEEE Conference on Computer Vision and
  Pattern Recognition}, pages 8601--8610, 2018.

\bibitem{sattler2012image}
Torsten Sattler, Tobias Weyand, Bastian Leibe, and Leif Kobbelt.
\newblock Image retrieval for image-based localization revisited.
\newblock In {\em BMVC}, volume~1, page~4, 2012.

\bibitem{quadnet}
Nikolay Savinov, Akihito Seki, Lubor Ladicky, Torsten Sattler, and Marc
  Pollefeys.
\newblock Quad-networks: unsupervised learning to rank for interest point
  detection.
\newblock In {\em Proceedings of the IEEE conference on computer vision and
  pattern recognition}, pages 1822--1830, 2017.

\bibitem{schonberger2016sfm}
Johannes~L Schonberger and Jan-Michael Frahm.
\newblock Structure-from-motion revisited.
\newblock In {\em Proceedings of the IEEE conference on computer vision and
  pattern recognition}, pages 4104--4113, 2016.

\bibitem{schonberger2017comparative}
Johannes~L Schonberger, Hans Hardmeier, Torsten Sattler, and Marc Pollefeys.
\newblock Comparative evaluation of hand-crafted and learned local features.
\newblock In {\em Proceedings of the IEEE conference on computer vision and
  pattern recognition}, pages 1482--1491, 2017.

\bibitem{disk}
Micha{\l}~J Tyszkiewicz, Pascal Fua, and Eduard Trulls.
\newblock {DISK}: Learning local features with policy gradient.
\newblock {\em arXiv preprint arXiv:2006.13566}, 2020.

\bibitem{van2020survey}
Jesper~E Van~Engelen and Holger~H Hoos.
\newblock A survey on semi-supervised learning.
\newblock {\em Machine Learning}, 109(2):373--440, 2020.

\bibitem{tilde}
Yannick Verdie, Kwang Yi, Pascal Fua, and Vincent Lepetit.
\newblock {TILDE:} a temporally invariant learned detector.
\newblock In {\em Proceedings of the IEEE Conference on Computer Vision and
  Pattern Recognition}, pages 5279--5288, 2015.

\bibitem{wilson_eccv2014_1dsfm}
Kyle Wilson and Noah Snavely.
\newblock Robust global translations with 1dsfm.
\newblock In {\em European conference on computer vision}, pages 61--75.
  Springer, 2014.

\bibitem{blendedmvs}
Yao Yao, Zixin Luo, Shiwei Li, Jingyang Zhang, Yufan Ren, Lei Zhou, Tian Fang,
  and Long Quan.
\newblock Blendedmvs: A large-scale dataset for generalized multi-view stereo
  networks.
\newblock In {\em Proceedings of the IEEE/CVF Conference on Computer Vision and
  Pattern Recognition}, pages 1790--1799, 2020.

\bibitem{lift}
Kwang~Moo Yi, Eduard Trulls, Vincent Lepetit, and Pascal Fua.
\newblock {LIFT:} learned invariant feature transform.
\newblock In {\em European Conference on Computer Vision}, pages 467--483.
  Springer, 2016.

\end{thebibliography}
}

\end{document}